\documentclass[letterpaper, 10 pt, conference]{ieeeconf}  

\IEEEoverridecommandlockouts                              

\usepackage{graphics} 
\usepackage{graphicx}
\usepackage{epsfig} 
\usepackage{times} 
\usepackage{amsmath} 
\usepackage{amssymb}  
\usepackage[linesnumbered,ruled,vlined]{algorithm2e}
\usepackage{url}
\usepackage{threeparttable}
\usepackage{multirow}
\usepackage{dsfont}
\usepackage{makecell}
\usepackage{booktabs}
\usepackage{soul, color, xcolor}
\soulregister\cite7
\soulregister\ref7 
\usepackage{bm}

\usepackage{cleveref}
\Crefname{figure}{Fig.}{Figs.}
\usepackage{numerica}

\usepackage[nolist]{acronym}

\newcommand{\highlight}[1]{\textcolor{black}{#1}}

\title{\LARGE \bf
ViTac-Tracing: Visual-Tactile Imitation Learning of Deformable Object Tracing
}

\author{Yongqiang Zhao$^1$*, Haining Luo$^2$*, Yupeng Wang$^1$, Emmanouil Spyrakos Papastavridis$^1$, \\Yiannis Demiris$^2$, Shan Luo$^1$ 
\thanks{$^1$Robot Perception Laboratory, Department of Engineering, King’s College London, Strand, London,
WC2R 2LS, United Kingdom, \{yongqiang.zhao, yupeng.1.wang, emmanouil.spyrakos, shan.luo\}@kcl.ac.uk.}
\thanks{ $^2$Personal Robotics Laboratory, Department of Electrical and Electronic Engineering, Imperial College London, London SW7 2AZ, United Kingdom \{haining.luo18, y.demiris\}@imperial.ac.uk.
}
\thanks{* Yongqiang Zhao and Haining Luo contributed equally to this work.}
}

\begin{document}

\begin{acronym}
    \acro{ee}[EE]{End-Effector}
    \acro{act}[ACT]{Action Chunking Transformer}
    \acro{dof}[DoF]{Degrees of Freedom}
    \acro{il}[IL]{Imitation Learning}
    \acro{bc}[BC]{Behavior Cloning}
    \acro{rl}[RL]{Reinforcement Learning}
    \acro{mpc}[MPC]{Model Predictive Control}
    \acro{sac}[SAC]{Soft Actor-Critic}
    \acro{1d}[1D]{one-dimensional}
    \acro{2d}[2D]{two-dimensional}
    \acro{kl}[KL]{Kullback–Leibler}
    \acro{mlp}[MLP]{Multilayer Perceptron}
    \acro{cnn}[CNNs]{Convolutional Neural Networks}
    \acro{ar}[AR]{Augmented Reality }
\end{acronym}

\maketitle
\thispagestyle{empty}
\pagestyle{empty}

\begin{abstract}
Deformable objects often appear in unstructured configurations. Tracing deformable objects helps bringing them into extended states and facilitating the downstream manipulation tasks. Due to the requirements for object-specific modeling or sim-to-real transfer, existing tracing methods either lack generalizability across different categories of deformable objects or struggle to complete tasks reliably in the real world. To address this, we propose a novel visual-tactile imitation learning method to achieve \ac{1d} and \ac{2d} deformable object tracing with a unified model. Our method is designed from both local and global perspectives based on visual and tactile sensing. Locally, we introduce a weighted loss that emphasizes actions maintaining contact near the center of the tactile image, improving fine-grained adjustment. Globally, we propose a tracing task loss that helps the policy to regulate task progression. On the hardware side, to compensate for the limited features extracted from visual information, we integrate tactile sensing into a low-cost teleoperation system considering both the teleoperator and the robot. Extensive ablation and comparative experiments on diverse \ac{1d} and \ac{2d} deformable objects demonstrate the effectiveness of our approach, achieving an average success rate of 80\% on seen objects and 65\% on unseen objects. Demos, code and datasets are available at \url{https://sites.google.com/view/vitac-tracing}.
\end{abstract}


\section{INTRODUCTION}
\begin{figure}[t]
	\centering
	\includegraphics[scale=0.34]{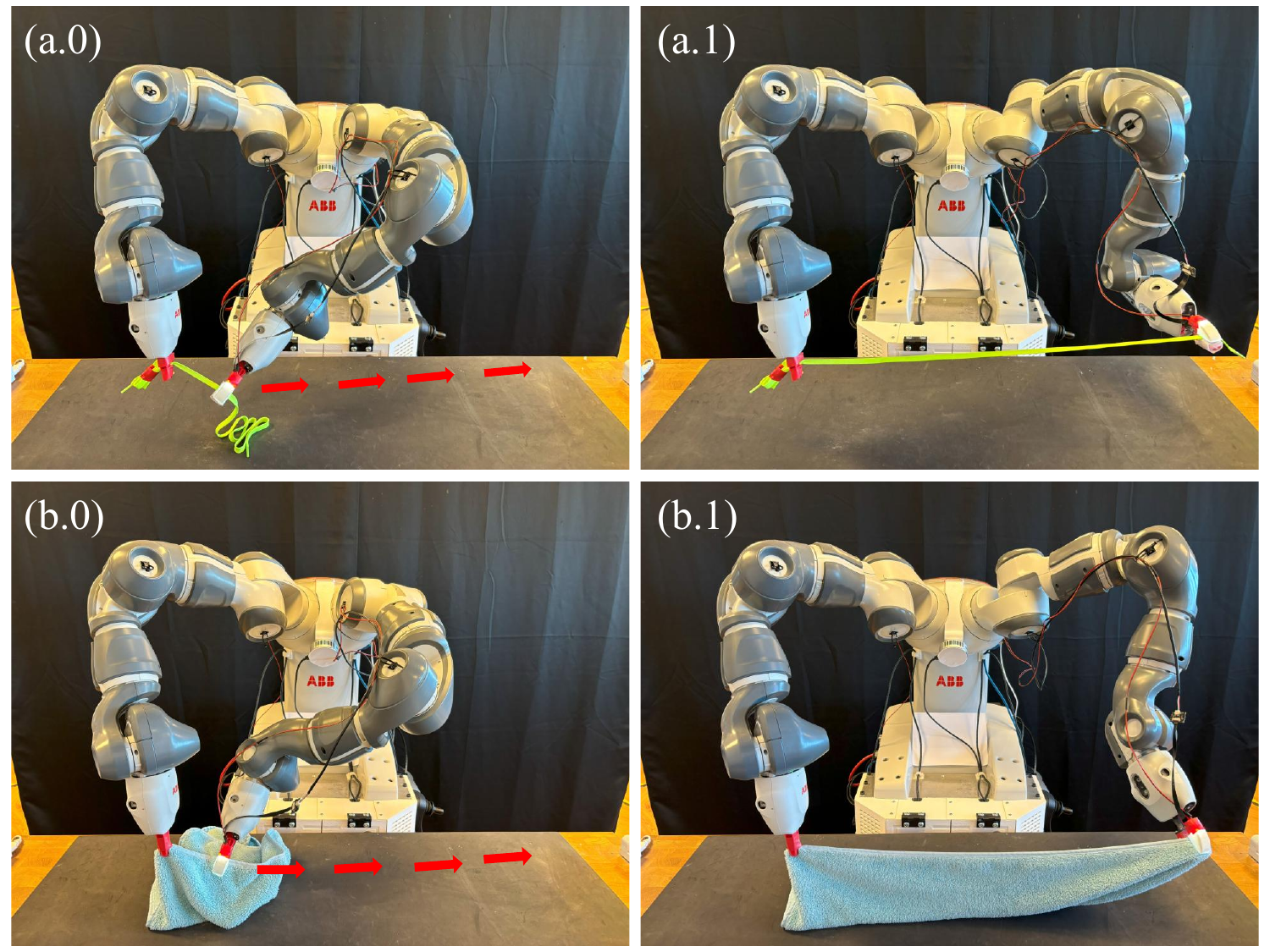}
        \caption{(a) \ac{1d} deformable object tracing. (b) \ac{2d} deformable object tracing. Using our proposed method, an ABB YuMi traces the objects by sliding a gripper along the \ac{1d} deformable object or an edge of the \ac{2d} deformable object and transforming them from the unstructured configurations on the left to the extended states on the right.}
        \label{fig:teaser}
	\vspace{-10pt}
\end{figure}

Deformable object manipulation has gained increasing attention in recent years due to its numerous real-world applications, such as cable management, cloth handling, and assistive dressing~\cite{zhu2022challenges}. A common characteristic of deformable objects is that they commonly appear in unstructured configurations, e.g. a folded-over shoelace or a crumpled towel, where their geometry and task-relevant features are difficult to observe directly (\Cref{fig:teaser}). Tracing transforms deformable objects into an extended configuration by following their edge with fingers from one end to the other ~\cite{she2021cable, sunil2023visuotactile}.

We consider deformable object tracing in two categories: \acf{1d} linear objects (e.g., cables, ropes) and \acf{2d} planar objects (e.g., towels, clothes). Although structurally distinct, these objects exhibit comparable physical properties~\cite{yin2021modeling}, suggesting the possibility of tracing both categories with a single unified model.
Previous work has so far addressed only one category, constrained by the limitations of their methods. Model-based controllers were among the first explored for the tracing task~\cite{she2021cable, zhang2023visual}. However, the requirement for accurate modeling of the object states and dynamics makes it difficult to construct a single controller for multiple objects. This is further compounded by the effectively infinite \ac{dof} of deformable objects.
\ac{rl} provides an alternative for learning unified tracing policies across multiple object types. However, it typically requires carefully designed reward functions and accurate deformable object modeling in simulation, and often suffers sim-to-real gaps~\cite{pecyna2022visual, sun2023learning}.
In contrast, \ac{il} methods leverage expert demonstrations to accelerate policy learning and avoids the need for explicit modeling or sim-to-real transfer. In this work, we propose an \ac{il} approach to trace both \ac{1d} and \ac{2d} deformable objects within a unified model.

During tracing, maintaining the contact region near the center of the fingertips helps minimize the risk of object slippage~\cite{she2021cable}. However, due to occlusion by the fingers, visual information alone is typically insufficient to capture grasping conditions. A common solution is to incorporate tactile sensing to provide the missing local information~\cite{she2021cable}. To date, most budget-efficient teleoperation systems do not include tactile sensing capabilities~\cite{zhao2023learning, wu2024gello, luo2025interface}. 

In this work, we utilize a dual-arm ABB YuMi and a bespoke 3D-printed kinematic model for teleoperation. Tactile sensing is incorporated by integrating a vision-based tactile sensor into the gripper finger. Previous studies have shown that sufficient visual and haptic feedback enhances the quality of demonstration data~\cite{xue2025reactive}. To this end, we provide the teleoperator with multi-modal feedback through real-time visual and tactile image streaming, complemented by vibration alerts for singularities.

With the visual and tactile information, we propose an imitation learning method for deformable object tracing. This work marks the first step towards a unified model for tracing both \ac{1d} and \ac{2d} deformable objects.
Our method is designed from both the local and global perspectives of the task. To improve fine-grained action adjustment and reduce the risk of dropping the object, we propose a local center loss that prioritizes actions that center the object in the tactile image.
Since various deformable manipulation tasks require tracing stopping at vision-detected accurate locations, e.g., stopping when inserting a cable into a clip during cable routing~\cite{luo2024multistage}, we further design a global task loss to regulate the task progression. 
Experiments validate the effectiveness of individual components and demonstrate the generalizability of our method, achieving an average success rate of 80\% on seen objects and 65\% on unseen objects.

In summary, the contributions of this work are as follows:
\begin{itemize}
    \item We propose a novel visual-tactile imitation learning framework for deformable object tracing, enabling real robots to trace on diverse \ac{1d} and \ac{2d} deformable objects through a unified policy;
    \item We introduce a budget-efficient visual–tactile teleoperation system with multi-modal feedback to enrich perception of both the robot and the teleoperator;
    \item Extensive experiments validate the effectiveness of the individual components of our method and demonstrate the performance on seen objects as well as the generalizability to unseen objects.
\end{itemize}

\section{RELATED WORKS}
\subsection{Deformable Object Tracing}
Deformable object manipulation has drawn increasing attention in robotics~\cite{yin2021modeling}. Progress has been made on tasks such as cable routing~\cite{lee2024sim}, knot tying~\cite{suzuki2021air}, rope untangling~\cite{viswanath2022autonomously, huang2023untangling}, and cloth folding~\cite{doumanoglou2016folding, lips2024learning}. A common facilitating step in these tasks is transforming the objects into extended or unfolded configurations. For example, spreading a T-shirt into a flat state can simplify the identification of keypoints or landmarks~\cite{canberk2022cloth}. Object tracing provides a means of transforming deformable objects into such states, and has been employed in some works on cloth unfolding~\cite{zhang2023visual} and cable insertion~\cite{she2021cable}.

In this work, we address the tracing task for both \ac{1d} and \ac{2d} deformable objects, with the aim of recovering fully extended configurations from unstructured, crumpled states. Owing to the effectively infinite number of \ac{dof} in deformable objects, tracing demands fine-grained adjustments to maintain stable gripper-object contact throughout the process. In recent years, several studies have attempted to tackle the tracing task for \ac{1d} linear objects (some referred to as “object following”) and for \ac{2d} planar objects (also known as “object sliding”).

Traditional control-based methods were among the first to be applied to the tracing task. For \ac{1d} objects, She et al.~\cite{she2021cable, sunil2023visuotactile} designed grip and pose controllers for cable tracing using real-time tactile feedback. Later, Liu et al.~\cite{liu2025dynamic} introduced a dynamic contrastive Koopman teleoperator for transparent tube tracing. For \ac{2d} objects, Zhang et al.~\cite{zhang2023visual} employed a deep \ac{mpc} method to iteratively servo garments from raw visual and tactile inputs. Although these approaches demonstrate promising results, they are tailored to specific object types and struggle to generalize across different geometries due to their reliance on accurate modeling of object dynamics and states.

\acf{rl} methods have also been applied to the tracing task \cite{pecyna2022visual, zhao2023skill, zhao2024fots, bi2025interactive}. Pecyna et al.~\cite{pecyna2022visual} achieved \ac{1d} object tracing in simulation using a model-free \ac{rl} approach with visual–tactile fusion. Later, Sun et al.~\cite{sun2023learning} extended the method to real \ac{1d} cable following using the \ac{sac} algorithm. Tailored for high-precision manipulation, Yu et al.~\cite{yu2023precise} applied a similar strategy to \ac{1d} needle threading. For \ac{2d} objects, Zhou et al.~\cite{zhou2021plas} trained an \ac{rl} policy in a latent action space for towel tracing. Although model-free \ac{rl} methods do not directly require object dynamics, these methods depend on either accurate deformable object models in simulation or carefully designed reward functions, which leads to sim-to-real gaps and hinders the development of the \ac{rl} environment for tracing both \ac{1d} and \ac{2d} objects.



As a highly effective robot manipulation method~\cite{fang2019survey}, \acf{il} offers another option for the tracing task. By leveraging expert demonstrations, it does not require accurate deformable object modeling and is not affected by sim-to-real gaps. This makes it an attractive choice for training a unified model capable of tracing multiple types of objects. In this study, we introduce a local center loss and a global task loss to an imitation learning framework, enabling us to address both \ac{1d} and \ac{2d} deformable object tracing.

\subsection{Vision-Tactile Imitation Learning}
\label{sec:vitac_bg}
In recent years, vision has become the primary modality used in \ac{il}~\cite{zhao2023learning, chi2023diffusion}. While visual sensing is well-suited for capturing global features, it is susceptible to occlusion by the manipulated object or surrounding items. Tactile sensing complements vision by providing rich local feature information.

Most existing teleoperation systems lack fine-grained tactile feedback~\cite{xue2025reactive}. Several recent studies have combined vision and tactile sensing for policy learning with \ac{il}. Xue et al.~\cite{xue2025reactive} developed a teleoperation system via \ac{ar} for demonstration collection and proposed a visual-tactile imitation learning method for contact-rich manipulation skills. Meanwhile, Gu et al.~\cite{gu2025tactilealoha} integrated a vision-based tactile sensor into the ALOHA system~\cite{zhao2023learning} to accomplish precise tasks such as zip-tie insertion and Velcro fastening. In this work, we take a similar approach and integrate a vision-based tactile sensor into a bimanual teleportation system.

In \ac{il}, collecting expert demonstrations is a crucial step. Existing data collection systems generally rely only on visual feedback~\cite{zhao2023learning, wu2024gello, luo2025interface}. While equipping robots with tactile sensing is straightforward, providing haptic feedback to human operators is equally important. Recent studies~\cite{gu2025tactilealoha, xue2025reactive}, employ real-time tactile image monitoring for this purpose, while commercial devices such as Touch Haptic Device~\cite{3dsystems2025touch} can also be used to provide vibration-based feedback. Sufficient visual and haptic feedback to the teleoperator can enhance the consistency and quality of demonstration data~\cite{xue2025reactive}. To enhance our data quality, we provide multi-modal feedback to the teleoperator with real-time visual–tactile monitoring and vibration feedback.

\section{PROBLEM STATEMENT}
\label{SEC:3}
We define deformable object tracing as the task of sliding a gripper along a \ac{1d} deformable object or an edge of a \ac{2d} deformable object, while maintaining continuous contact, until reaching the terminal point. We assume that one end of the object is secured by a second gripper, and the sliding gripper starts near the grasping point and moves towards the free end of the object. The task is illustrated in \Cref{fig:teaser}.

The deformable object tracing problem can be formulated as follows. We model a \ac{1d} deformable object, or an edge of a \ac{2d} deformable object, as a time-varying spatial curve $\mathcal{C}_t \subset \mathbb{R}^3$ in Cartesian space. The length of $\mathcal{C}_t$ is defined as $L$. The time steps of this task move from $t=0$ to $t=T$. Let $p_0=(x_0,y_0,z_0)$ denote the fixed point in the world frame, $p_t=(x_t,y_t,z_t)$ is the contact point with the moving gripper. $o^T$ represents the gripper's tactile sensing region. 

In the deformable object tracing task, we identify two constraints for successful completion: (1) the object should remain grasped, i.e., $p_t \in \mathcal{C}_t$, meaning the contact point $p_t$ should always lie on the curve; (2) the contact point $p_t$ should stay within the effective tactile sensing region $o^T$, i.e., $p_t \in o^T$. Since the tactile sensing region covers the fingertip, leaving this region implies the objects has slipped or dropped. Meanwhile, we define two objectives for the task: (1) the distance between the contact point $p_t$ and the fixed point $p_0$ should gradually converge to the total curve length as $t$ increases, i.e., $||p_t-p_0||_2 \rightarrow L$; (2) this distance function between should be monotonically increasing over time, i.e., $\frac{d}{dt}||p_t-p_0||_2\geq 0$.

\section{METHODOLOGY}
\label{SEC:4}
\subsection{Overview}
\label{SEC:4-1}
\begin{figure}[t] 
    \centering
    \includegraphics[width=1.0\linewidth]{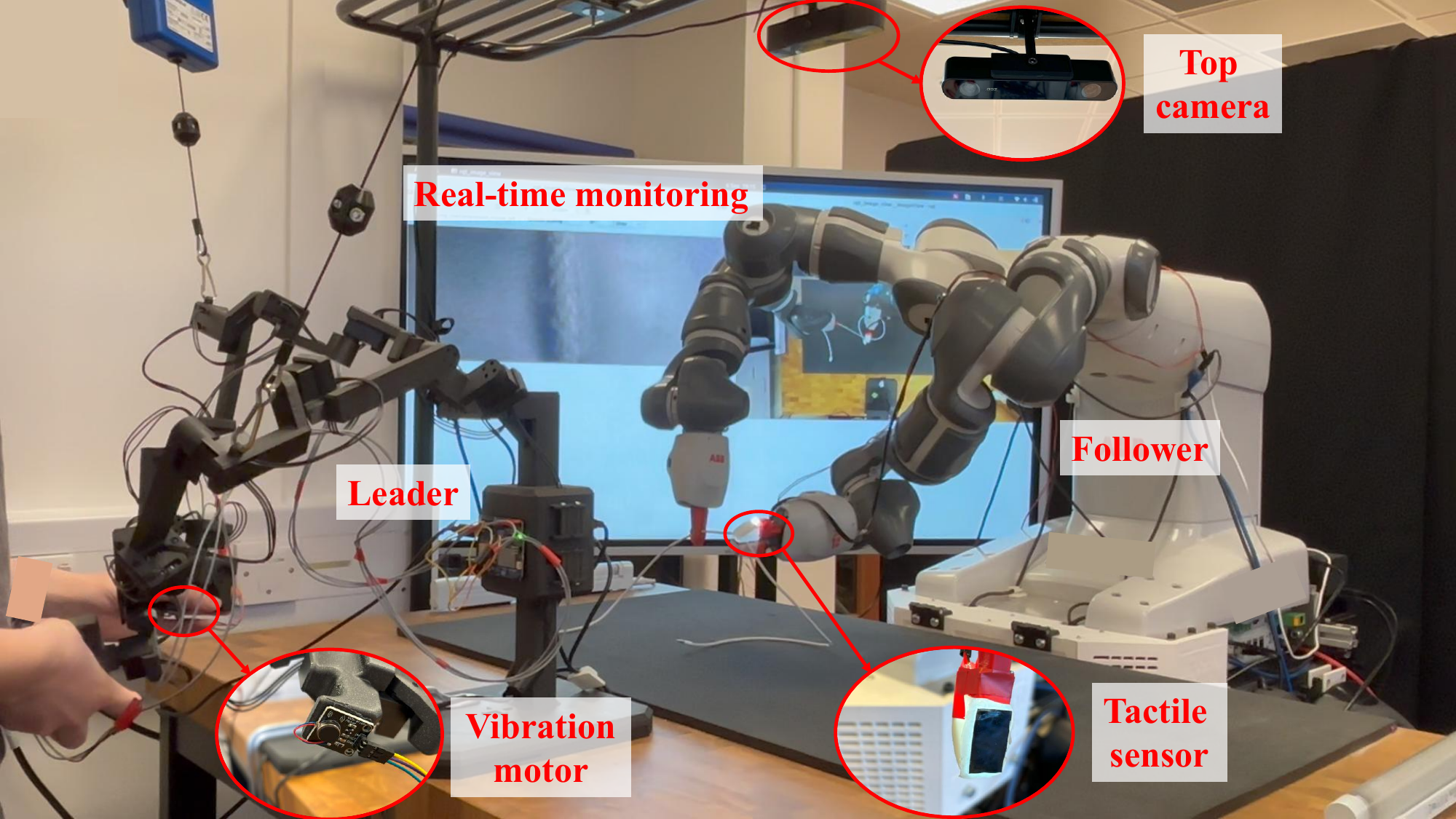}
    \caption{Visual-tactile teleoperation system for collecting demonstrations. On the robot side, a top-view camera is installed, and a tactile sensor is mounted on the follower robot’s gripper. On the teleoperator side, visual and tactile images are monitored in real time, and vibration motors are mounted on the leader robot’s gripper.}
    \label{fig:tagumi}
\end{figure}

The proposed visual-tactile imitation learning framework is illustrated in \Cref{fig:framework}. Demonstration data is collected using a visual-tactile teleoperation system equipped with real-time visual/tactile monitoring and vibration feedback, as shown in \Cref{fig:tagumi}. We then use the collected dataset to train the proposed action-chunking-based policy with the local center loss, the \ac{kl} divergence loss, and the global task loss. The trained policy is subsequently deployed on a real robot.

\subsection{Visual-Tactile Teleoperation System}
\label{SEC:4-2}

\subsubsection{\textbf{Robot Side}} As discussed in \Cref{sec:vitac_bg}, tactile sensors can provide rich information about grasping conditions. A vision-based tactile sensor, derived from GelSight Wedge~\cite{wang2021gelsight}, is mounted on the gripper. For the visual input, a ZED 2 stereo camera provides a top-down view of the Cartesian space (\Cref{fig:tagumi}). The robot is controlled in joint position mode, with the \ac{ee} velocity limited to 400 mm/s. The robot platform is further equipped with an Nvidia Jetson Orin running the camera driver and the robot's ROS driver. All drivers are containerized in Docker on ROS Noetic.

\subsubsection{\textbf{Teleoperator Side}} \Cref{sec:vitac_bg} also highlighted the importance of haptic feedback. The absence of such feedback can limit the teleoperator’s awareness of grasping conditions. To address this, we enhance the teleoperator's perception by adding a display screen to stream visual and tactile image from the robot side in real time. Meanwhile, as our tracing policy is trained with \ac{ee} poses rather than joint angles as in~\cite{wang2021gelsight} (see \Cref{sec:proprioception_experiment} for the rationale behind this choice), singular configurations become a concern due to the non-unique mapping between \ac{ee} poses to joint configurations. To address this, we provide haptic feedback through vibration motors (DAOKAI DC 5V Mini) mounted on the leader grippers, which are triggered when the \ac{ee} approaches singular configurations. To calculate the proximity to singularity, we adopt Yoshikawa Manipulability Index~\cite{yoshikawa1985manipulability}

\begin{equation}
    w(q) = \sqrt{\mathrm{det}(J(q)J(q)^T)},
\end{equation}
where $q$ is the joint states of the robot, and $J(\cdot)$ is the Jacobian matrix. Higher $w(q)$ indicates better dexterity, meaning the robot can move in multiple directions from the current configuration. \highlight{The motors are triggered when the index falls below $\lambda_w\mathrm{max}(w)$, where $\lambda_w=0.2$ is selected via grid search in our experiments.} All three feedback devices are shown in \Cref{fig:tagumi}.

Using this visual-tactile teleoperation system, we collect kinematic information $o^K$, visual information $o^V$, and tactile information $o^T$ from the robot side, while ground-truth action information $a$ is collected from the teleoperator side. We define the observation at time step $t$ as $o_t=\{o_t^K, o_t^V,o_t^T\}$. A collected demonstration episode $\mathcal{D}$ is thus represented as $\mathcal{D} = \{(o_t, a_t)\}_{t=0}^T$.

\subsection{Tracing Policy Learning}
\label{SEC:4-3}
\begin{figure*}[ht]
	\centering
	\includegraphics[scale=0.4]{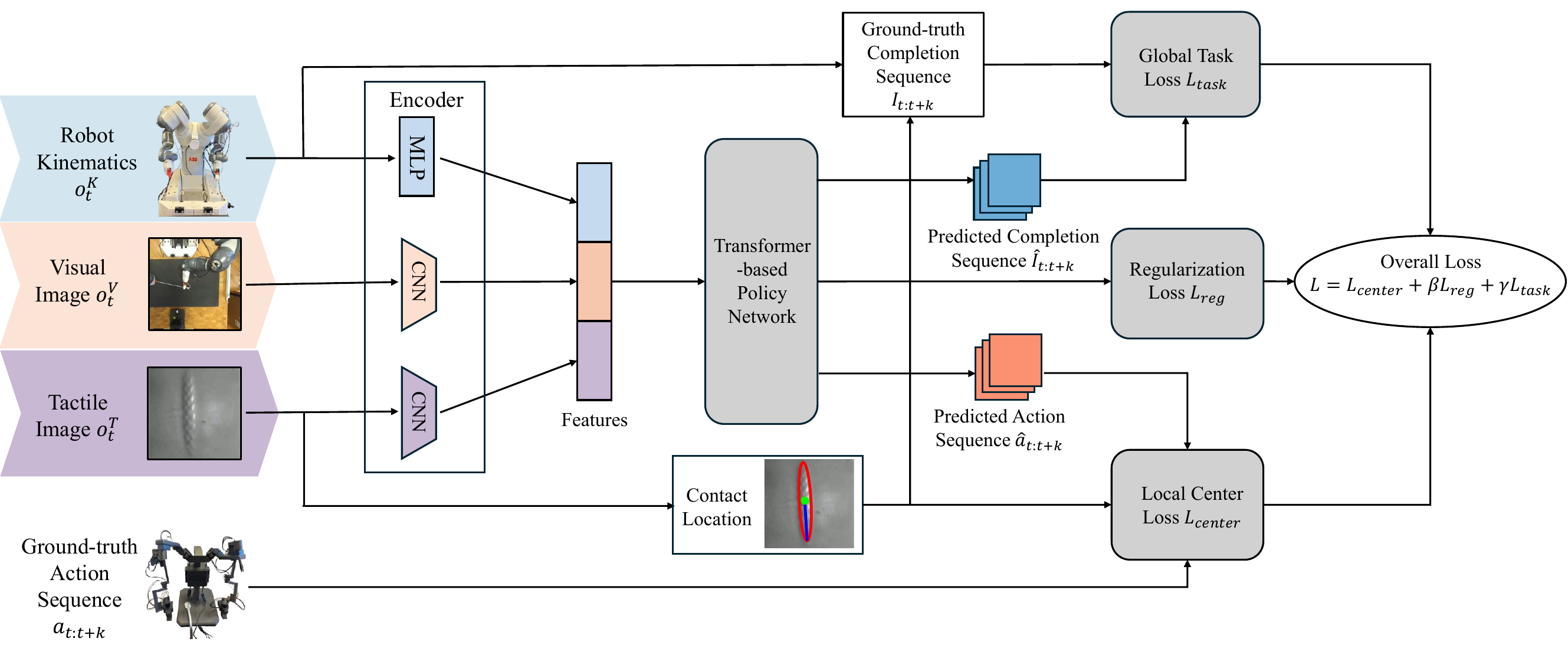}
        \caption{Overview of the proposed tracing policy learning framework. The inputs include robot kinematics $o^K_t$, visual image $o^V_t$, and tactile image $o^T_t$ collected from the follower robot, while the ground truth consists of action sequence $a_{t:t+k}$ recorded from the leader robot. Input features are first extracted using a \acf{mlp} and \acf{cnn}. These features are then concatenated and fed into a Transformer-based policy network, which is trained using a combination of three loss functions: local center loss, global task loss, and regularization loss.}
        \label{fig:framework}
	\vspace{-0.4cm}
\end{figure*}

We train the tracing policy via transformer-based policy network, as illustrated in \Cref{fig:framework}. Specifically, we adopt the \ac{act} algorithm~\cite{zhao2023learning}, where the policy $\pi_\theta(\hat{a}_{t:t+k}|o_t)$ takes an observation $o_t$ as input and predicts the next $k$ sequential actions $\hat{a}$. \ac{act} optimizes a reconstruction loss $\mathcal{L}_{reconst}$ to reconstruct actions from the inputs and a regularization loss $\mathcal{L}_{reg}$ to regularize the Transformer encoder toward a Gaussian prior. 

\begin{equation}
    \mathcal{L}_{reconst}=\mathrm{MAE}(\hat{a}_{t:t+k},a_{t:t+k}),
\end{equation}
\begin{equation}
    \mathcal{L}_{reg}=D_{KL}(q_{\phi}(z|a_{t:t+k}, \overline{o}_t)|| \mathcal{N} (0, I )),
\end{equation}
where $\mathrm{MAE}$ is the mean absolute error, i.e., L1 loss. $q_{\phi}(z|a_{t:t+k}, \overline{o}_t)$ denotes the Transformer encoder, $z$ is a style variable, and $\overline{o}_t$ represents $o_t$ without image observations. We refer the reader to \cite{zhao2023learning} for the technical details of \ac{act}.

In addition to the aforementioned losses, we introduce a local center loss to enhance tactile-feedback-driven adjustments, and a global task loss to improve task progression.

\subsubsection{\textbf{Local Center Loss}}

\begin{figure}[t] 
    \centering
    \includegraphics[width=1.0\linewidth, trim=0.0cm 0.0cm 0.0cm 0.0cm, clip]{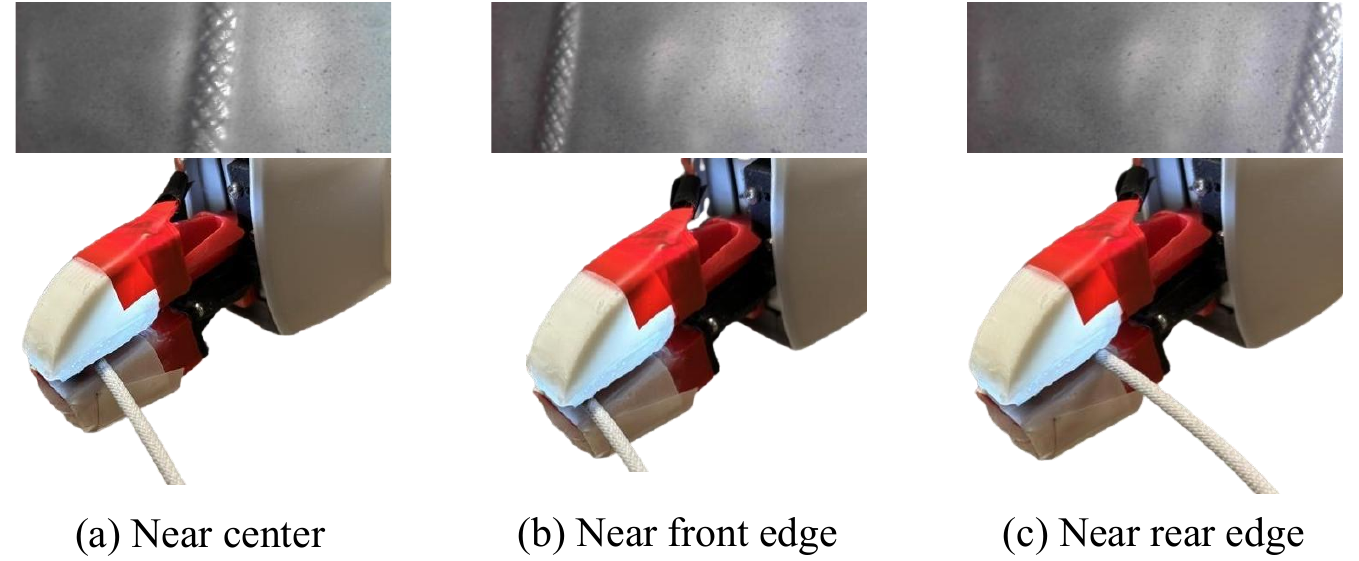}
    \caption{Tactile images under different contact locations between the object and gripper. (a) Object grasped near the center of the tactile sensing region; (b) Object grasped near the front edge of the tactile sensing region; (c) Object grasped near the rear edge of the tactile sensing region. When the object is grasped near the edges of the sensing region, it is more likely to slip into an imperceptible area.}
    \label{fig:center_loss}
    \vspace{-0.4cm}
\end{figure}

As noted in Section \ref{SEC:3}, a key constraint in the tracing task is maintaining contact within the gripper’s tactile sensing region, i.e., $p_t\in o^T$. Real-time tactile monitoring allows teleoperators to perceive gripper-object contact during demonstration collection. However, owing to the high-\ac{dof} dynamics of deformable objects, contacts near the edges of the tactile sensing region are prone to drop from the robot gripper or slip to the imperceptible area, as illustrated in \Cref{fig:center_loss}. The ideal contact location lies at the center of the sensing region. To encourage this, we introduce a center loss that prioritizes the actions leading to contact points moving closer to the sensor center.

\highlight{We first localize the contact point $p_t$ via classical image processing, since the high-resolution tactile texture is highly salient, enabling robust contact-point extraction, as shown in \Cref{fig:objects_tactile}.} A contact mask is generated through grayscale conversion, thresholding, and Gaussian filtering, followed by contour extraction. The largest contour is then fitted with an ellipse (preferred) or characterized via Principal Component Analysis (PCA), yielding the pixel-coordinate contact point $p_t^{\mathrm{tac}}=(u_t^{\mathrm{tac}},v_t^{\mathrm{tac}})$. Let the sensing-region center be $c=(u_c,v_c)$, we introduce a weight factor
\begin{equation}
    \highlight{w_t = exp(-\frac{||p^{tac}_t-c||}{c}).}
\end{equation}

We combine the original reconstruction loss with the proposed weight to obtain the local center loss

\begin{equation}
    \mathcal{L}_{center} = w_{t:t+k}\cdot\mathrm{MAE}(\hat{a}_{t:t+k},a_{t:t+k}).
\end{equation}

\subsubsection{\textbf{Global Task Loss}}
The definition of tracing discussed in \Cref{SEC:3} illustrates the constraint of maintaining object–gripper contact and the goal of reaching the full curve length. While the center loss addresses the contact constraint, it does not assist the policy in achieving the overall task goal. In many manipulation tasks, precise termination is critical, e.g., tracing to the corner of a towel to enable folding, or stopping before a cable slides off from the gripper in insertion tasks. We therefore introduce a global task loss to regulate the task progression.

\highlight{Assuming the object edge is locally straight during the task, where one end is fixed and the segment between the fixed point $p_0$ and the contact point $p_t$ is typically tensioned straight, the traced length can be inferred from $p_0$ and $p_t$, and the completion index $I$ is computed from the traced length and the overall length.} We firstly localize $p_t^{\mathrm{tac}}$ from the tactile image and convert it to the gripper frame (coincident with the sensor frame)
\begin{equation}
    p_t^{gripper} = (\frac{u_t^{tac}-u_c}{p2m}, \frac{v_t^{tac}-v_c}{p2m}, 0),
\end{equation}
where $p2m$ is the pixel-to-meter scale.

Then, we can calculate $p_t$ through $p_t^{tac}$ using coordinate transformation

\begin{equation}
    p_t = T_{gripper}^{world}p_t^{gripper},
\end{equation}
where $T_{gripper}^{world}$ is the transform matrix from the gripper frame to the world frame.

Following that, the ground-truth completion index $I$ of task is able to be labeled on the collected demonstrations as
\begin{equation}
    I = \mathrm{min}(\mathrm{max}(\frac{||p_t-p_0||_2}{||p_T-p_0||_2}, 0) , 1).
\end{equation}

As shown in \Cref{fig:framework}, we extend the policy network with a completion index prediction branch, which predicts completion squence $\hat{I}_{t:t+k}$ alongside the action sequence. This branch is optimized using
\begin{equation}
\mathcal{L}_{task} = \mathrm{MSE}(\hat{I}_{t:t+k}, I_{t:t+k}),
\end{equation}
where $\mathrm{MSE}$ denotes the mean squared error (L2 loss).

Therefore, the overall loss $\mathcal{L}$ of the proposed network is
\begin{equation}
   \highlight{ \mathcal{L} = \mathcal{L}_{center} + \lambda_{reg} \mathcal{L}_{reg} + \lambda_{task}\mathcal{L}_{task},}
\end{equation}
\highlight{where $\lambda_{reg}=100$ and $\lambda_{task}=100$ are selected via grid search.}

\subsection{Task Evaluation Metrics}
Following \cite{pecyna2022visual, zhang2023visual, luo2025interface}, we identify the experiment results and use their rates as evaluation metrics
\begin{itemize}
    \item \textbf{Success}: the gripper follows the object to the terminal end and maintains grasp there (any point within the final $5\%$ of length).
    \item \textbf{Robot collision}: the robot collides with the object or itself and cannot recover from the current state.
    \item \textbf{Early stopping}: the gripper does not reach the terminal end but does not drop the object.
    \item \textbf{Over-tracing}: the gripper reaches the final $5\%$ but fails to maintain grasp.
    \item \textbf{Object dropping}: the gripper drops the object before reaching the end.
\end{itemize}

We introduce two other metrics to further quantify the policy performance: 
\begin{itemize}
    \item \textbf{Success time}: the time length the robot spends to complete the tracing task successfully.
    \item \textbf{Completion ratio}: the ratio between the distance the gripper reached (measured this distance from the final contact point to the fixed point) and the deformable object length, i.e., $||p_T-p_0||_2/L$.
\end{itemize}

\section{EXPERIMENT RESULTS}
\subsection{Experimental and Training Setups}
\begin{figure}[t] 
    \centering
    \includegraphics[width=0.8\linewidth, trim=0.0cm 0.65cm 0.0cm 0.0cm, clip]{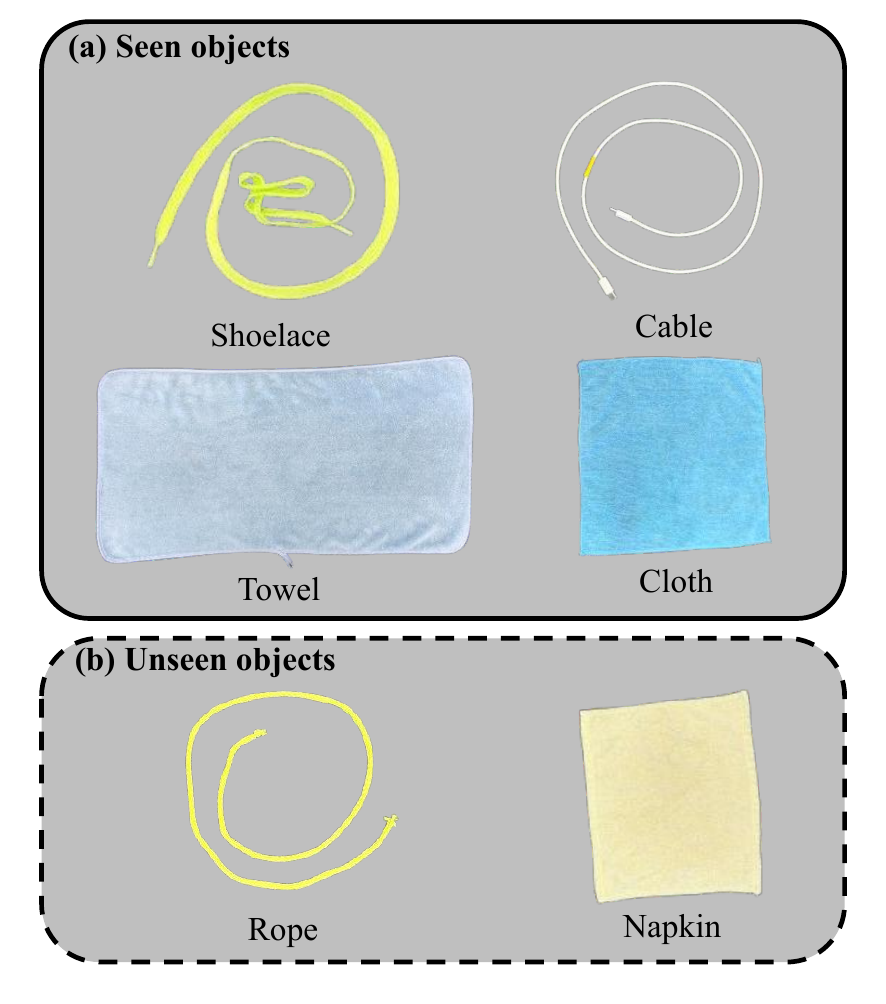}
    \caption{\ac{1d} and \ac{2d} deformable objects used in the experiments. (a) Seen objects; (b) Unseen objects. For each seen object, 25 demonstrations were collected. The trained models were tested on both seen and unseen objects.}
    \label{fig:objects}
    \vspace{-0.5cm}
\end{figure}

\begin{figure}[ht] 
    \centering
    \includegraphics[width=0.8\linewidth]{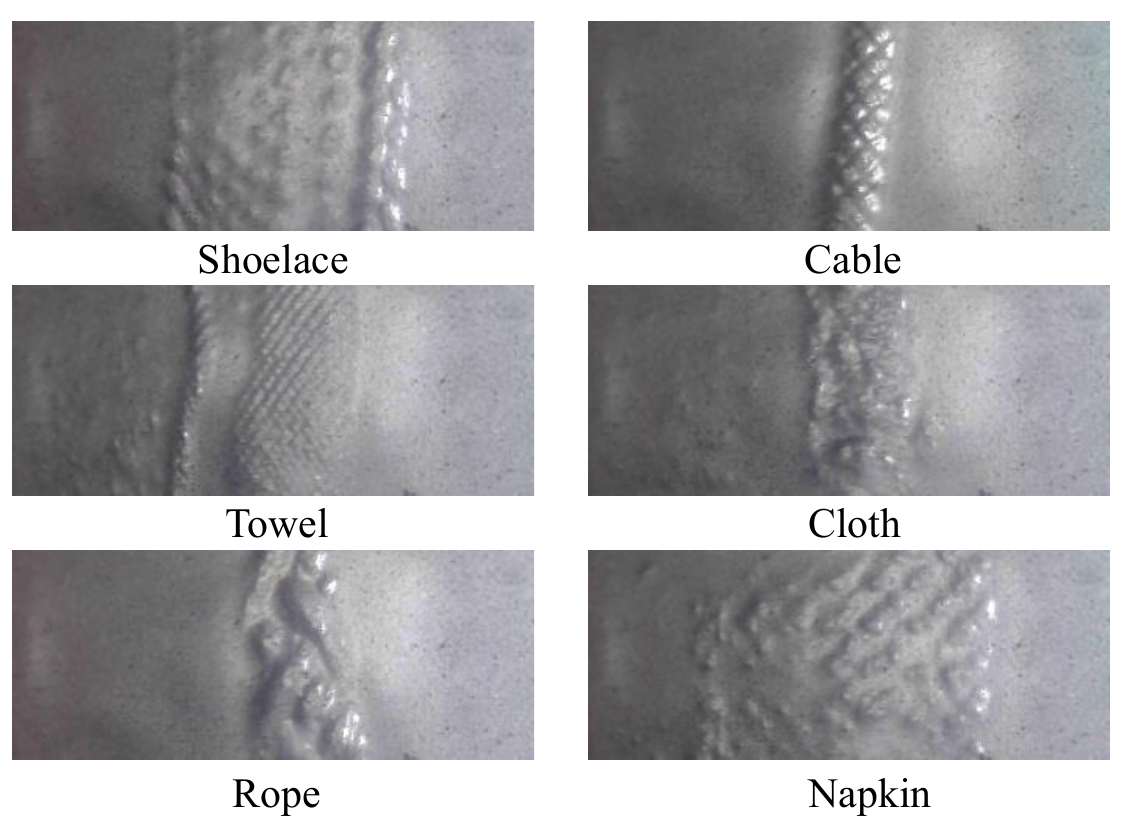}
    \caption{Tactile images providing high-resolution texture information of the tested \ac{1d} deformable objects and the hemmed edges of \ac{2d} deformable objects. The folded hems of the \ac{2d} objects show textures similar to the \ac{1d} objects. The tactile textures of the unseen objects also exhibit resemblance to those of the seen objects.}
    \label{fig:objects_tactile}
    \vspace{-0.8cm}
\end{figure}

For data collection, we selected two \ac{1d} and two \ac{2d} deformable objects: a flat shoelace, a braided cable, a face towel, and a microfiber cloth (\Cref{fig:objects}). We further show the example tactile images of the tested objects in \Cref{fig:objects_tactile}. For each object, 25 demonstrations were collected using our visual–tactile teleoperation system, resulting in 100 demonstrations in total. During collection, we recorded visual images, tactile images, robot states, and kinematic model states as action commands at 30~Hz. The visual and tactile images were cropped and resized to 480~×~480. The 16-dimensional action–state input consisted of either 14 joint states with 2 gripper states (for models trained in joint space) or 2 \ac{ee} poses with 2 gripper states (for models trained in Cartesian space). We further test our policies on two unseen objects: a \ac{1d} synthetic rope and a \ac{2d} cotton napkin. \highlight{In addition, we keep lighting constant, use a black sponge pad as the table surface, and randomize object placement by stacking the 1D and 2D deformable objects during data collection and policy deployment.}

We set the chunk size to 60 steps and use separate ResNet18 backbones to extract visual and tactile features. During training, image brightness, contrast, and gamma are randomized. Each model is trained for 15,000 epochs, and the checkpoint with the lowest validation loss is selected. Training is performed on a workstation equipped with an AMD Ryzen Threadripper Pro 5965WX processor, 128~GB of RAM, and an RTX~4090 GPU. Temporal aggregation is disabled, following \ac{act}’s tuning recommendations~\cite{luo2025interface}. During inference, the models run on a desktop computer with an Intel i9 CPU (24 cores), 32~GB of RAM. For models trained in Cartesian space, the output \ac{ee} poses are converted to joint angles before being executed.

\subsection{Proprioceptive Representation Comparison}
\label{sec:proprioception_experiment}
The two common representations for proprioception are the joint angles and \ac{ee} poses. While joint angles provide raw low-level proprioception, end-effector poses represent a higher-level kinematic abstraction. To investigate the effect of proprioception representation on the tracing task, we compared models trained with two different forms of input with all four datasets. 

\begin{figure}[t] 
    \centering
    \includegraphics[width=0.7\linewidth]{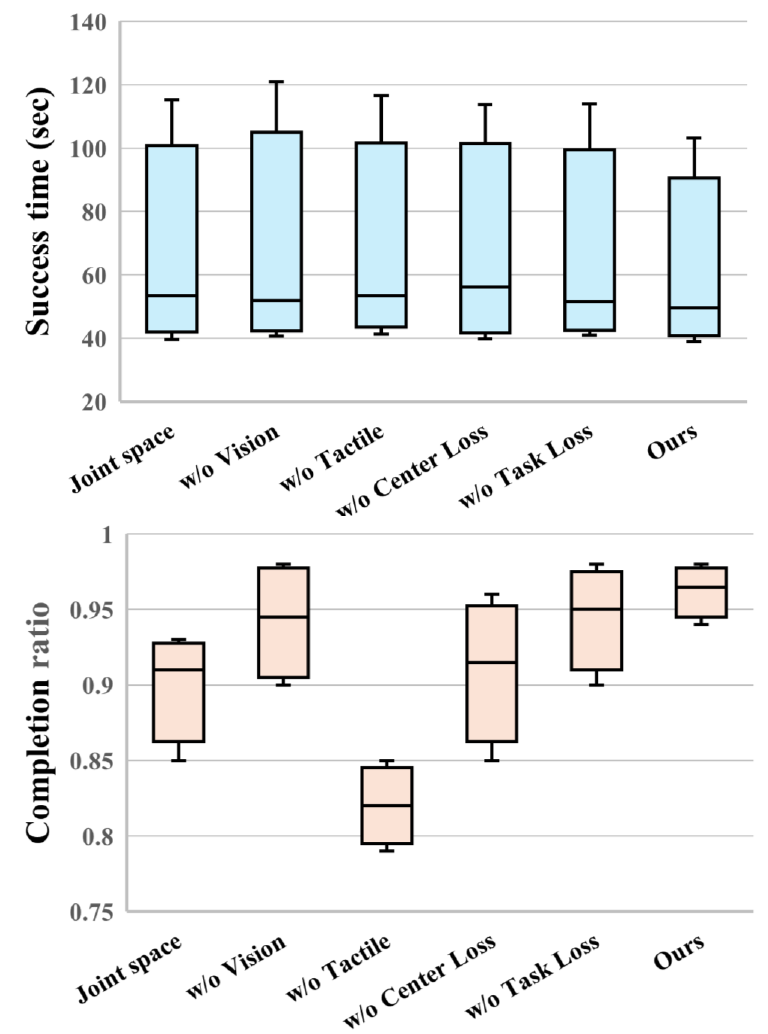}
    \caption{Success time and completion ratio for the component ablation study. Overall, the tested models showed similar success time. It is observable that all variants demonstrated lower completion ratios, indicating degraded task performance.}
    \label{fig:errbar}
    \vspace{-0.4cm}
\end{figure}

We evaluated both models on the four objects used for data collection, with 10 trials per object, resulting in 40 experiments per model. The accumulated results are shown in \Cref{Tab:1}. Compared with the joint-space model, the Cartesian-space model (last row) achieved a higher success rate of 80\%. This trend is also visible in \Cref{fig:errbar}, where the completion ratio of the joint-space model is noticeably lower. These findings suggest that the higher-level abstraction of \ac{ee} poses is more suitable for the tracing task. Since tracing is inherently defined in task space, providing the policy with \ac{ee} poses aligns the input with the task objective and avoids redundancy present in joint-angle representations. This reduces ambiguity and enables finer-grained adjustments, which is evident in the lower object-dropping rate. The joint-space model exhibits a much higher frequency of object drops, indicating a weaker ability to adjust object orientation relative to the fingers. Therefore, we adopt \ac{ee} poses as the proprioceptive representation for the remainder of this work.

\subsection{Component Ablation Study}
To evaluate the contribution of different sensing modalities and objective terms, we performed an ablation study. Specifically, we compared the proposed system against variants trained without tactile input, without visual input, without the proposed center loss, and without the proposed task loss. Each model was tested in 10 trials per object, and the results were aggregated (\Cref{Tab:1}). Overall, the results show that both visual and tactile information, as well as the two proposed losses, contributed to the success of the experiments. The success time across all tested models did not exhibit noticeable differences.

\begin{table}
\centering
\caption{Ablation experimental results to validate the effectiveness of individual components.}
\label{Tab:1}
\setlength{\tabcolsep}{0.3mm}{
\begin{tabular}{c|c|cccc}
\hline
\textbf{Methods} & \textbf{\begin{tabular}[c]{@{}c@{}}Success rate \\ (Wilson 95\% CI)\end{tabular}} & \textbf{\begin{tabular}[c]{@{}c@{}}Robot \\ collision\end{tabular}} & \textbf{\begin{tabular}[c]{@{}c@{}}Early \\ stopping\end{tabular}} & \textbf{\begin{tabular}[c]{@{}c@{}}Over- \\ tracing\end{tabular}} & \textbf{\begin{tabular}[c]{@{}c@{}}Object \\ dropping\end{tabular}} \\ \hline\hline
\textbf{Joint Space}      & 70.0\% [54.6, 81.9]                   & 1/40                        & 4/40                    & 2/40                          & 5/40 \\
\textbf{w/o Vision}       & 65.0\% [49.5, 77.9]                   & 4/40                        & 2/40                    & 8/40                          & 0/40 \\
\textbf{w/o Tactile}      & 60.0\% [44.6, 73.7]                   & 2/40                        & 5/40                    & 1/40                          & 8/40 \\
\textbf{w/o Center Loss}  & 65.0\% [49.5, 77.9]                   & 4/40                        & 1/40                    & 0/40                          & 9/40 \\
\textbf{w/o Task Loss}    & 67.5\% [52.0, 79.9]                   & 3/40                        & 3/40                    & 7/40                          & 0/40 \\
\textbf{Ours}             & \textbf{80.0\% [65.2, 89.5]}          & 2/40                        & 2/40                    & 3/40                          & 1/40 \\ \hline
\end{tabular}
}
\end{table}

\begin{table}
\centering
\caption{Experimental results using different training datasets.}
\label{Tab:2}
\setlength{\tabcolsep}{0.3mm}{
\begin{tabular}{c|c|c|cccc}
\hline
\textbf{\begin{tabular}[c]{@{}c@{}}Dataset \\ types\end{tabular}}& \textbf{\begin{tabular}[c]{@{}c@{}}Tested \\ objects\end{tabular}} & \textbf{\begin{tabular}[c]{@{}c@{}}Success rate\\ (Wilson 95\% CI)\end{tabular}} & \textbf{\begin{tabular}[c]{@{}c@{}}Robot \\ collision\end{tabular}} & \textbf{\begin{tabular}[c]{@{}c@{}}Early \\ stopping\end{tabular}} & \textbf{\begin{tabular}[c]{@{}c@{}}Over- \\ tracing\end{tabular}} & \textbf{\begin{tabular}[c]{@{}c@{}}Object \\ dropping\end{tabular}} \\ \hline\hline
\textbf{Shoelace}      & \textbf{\textbackslash{}}                                                         & 80.0\% [49.0, 94.3]                                                     & 1/10                                                      & 0/10                                                      & 1/10                                                      & 0/10                                                      \\
\textbf{Cable}         & \textbackslash{}                                                                  & \textbf{90.0\% [59.6, 98.2]}                                            & 1/10                                                      & 0/10                                                      & 0/10                                                      & 0/10                                                      \\
\textbf{Towel}         & \textbackslash{}                                                                  & 60.0\% [31.3, 83.2]                                                     & 2/10                                                      & 1/10                                                      & 1/10                                                      & 0/10                                                      \\
\textbf{Cloth}         & \textbackslash{}                                                                  & \textbf{80.0\% [49.0, 94.3]}                                            & 0/10                                                      & 1/10                                                      & 1/10                                                      & 0/10                                                      \\ \hline
\textbf{\ac{1d}}            & \textbf{\begin{tabular}[c]{@{}c@{}}Shoelace\\ Cable\end{tabular}}                 & \begin{tabular}[c]{@{}c@{}}\textbf{90.0\% [59.6, 98.2]} \\ 80.0\% [49.0, 94.3] \end{tabular}         & \begin{tabular}[c]{@{}c@{}}0/10 \\ 1/10 \end{tabular}         & \begin{tabular}[c]{@{}c@{}}0/10 \\ 0/10 \end{tabular}         & \begin{tabular}[c]{@{}c@{}}1/10 \\ 1/10 \end{tabular}         & \begin{tabular}[c]{@{}c@{}}0/10 \\ 0/10 \end{tabular}         \\ \hline
\textbf{\ac{2d}}            & \textbf{\begin{tabular}[c]{@{}c@{}}Towel\\ Cloth\end{tabular}}                    & \begin{tabular}[c]{@{}c@{}}\textbf{70.0\% [39.7, 89.2]} \\ 70.0\% [39.7, 89.2] \end{tabular}         & \begin{tabular}[c]{@{}c@{}}1/10 \\ 0/10 \end{tabular}         & \begin{tabular}[c]{@{}c@{}}1/10 \\ 1/10 \end{tabular}         & \begin{tabular}[c]{@{}c@{}}0/10 \\ 1/10 \end{tabular}         & \begin{tabular}[c]{@{}c@{}}1/10 \\ 1/10 \end{tabular}         \\ \hline
\textbf{Ours}          & \textbf{\begin{tabular}[c]{@{}c@{}}Shoelace\\ Cable\\ Towel\\ Cloth\end{tabular}} & \begin{tabular}[c]{@{}c@{}}\textbf{90.0\% [59.6, 98.2]} \\ 80.0\% [49.0, 94.3] \\ \textbf{70.0\% [39.7, 89.2]} \\ \textbf{80.0\% [49.0, 94.3]} \end{tabular} & \begin{tabular}[c]{@{}c@{}}0/10 \\ 1/10 \\ 1/10 \\ 0/10 \end{tabular} & \begin{tabular}[c]{@{}c@{}}1/10 \\ 1/10 \\ 0/10 \\ 0/10 \end{tabular} & \begin{tabular}[c]{@{}c@{}}0/10 \\ 0/10 \\ 1/10 \\ 0/10 \end{tabular} & \begin{tabular}[c]{@{}c@{}}0/10 \\ 0/10 \\ 1/10 \\ 2/10 \end{tabular} \\ \hline
\end{tabular}
}
\end{table}

\subsubsection{\textbf{Ablation on Sensing Modalities}}
We first assess the effect of removing tactile or visual input to determine the role of each sensing modality. Without visual information, the robot is more prone to over-tracing, reflecting its reduced ability to control task termination. In contrast, without tactile information, the robot achieves lower completion ratio (\Cref{fig:errbar}) and is more likely to drop the object, indicating weaker fine-grained adjustment of the tracing direction. These results suggest that the model relies more on visual information for task progression, while tactile information plays a key role in maintaining stable contact between the object and the gripper.

\subsubsection{\textbf{Ablation on Objective Losses}}
We also examine the impact of excluding the center loss and task loss to evaluate their contribution to policy performance. Without the center loss, the policy exhibits noticeably higher rates of object dropping. This suggests that the regulation introduced by the center loss helps the model learn adjustment actions more effectively. Although our demonstration data consist only of successful episodes, teleoperator actions are not always ideal, i.e., they do not consistently move the contact region towards the fingertip center. The center loss weighs actions and prioritizes those keep the object near the center of the tactile image, thereby improving the model’s learning of tracing direction adjustments. Without the task loss, failure modes show a higher rate of over-tracing and slightly more cases of early stopping. This indicates that including the completion index as an output enables the model to better track task progression and recognize the task endpoint.

\highlight{The dominant failure modes are: (1) robot collision due to excessive contact force and friction/entanglement; (2) early stopping/over-tracking under visual occlusion near the object end; and (3) object dropping due to weak tactile feedback and an unstable grasp. The towel shows a lower success rate mainly because its longer length amplifies accumulated tracking errors and interaction-induced disturbances.}

\subsection{Effect of Multi-Object Training}
The previous experiments have validated the design choices of our method. In this section, we investigate whether unifying deformable object tracing with a single model improves or harms task performance. We train models with individual object datasets (top four rows in \Cref{Tab:2}), with grouped \ac{1d} and \ac{2d} datasets (middle section of \Cref{Tab:2}), and with all object datasets combined (bottom section of \Cref{Tab:2}). Each model is then tested on its corresponding objects, e.g., the model trained on the shoelace dataset is tested on the shoelace. Overall, the success rate is higher for \ac{1d} objects and the results indicate similar performance between unified and individual models.

For the three models tested on the shoelace, the success rate ranges between 80\% and 90\%, and similar results are observed for the cable. For the towel, the success rate falls between 60\% and 70\%, while for the cloth it is between 70\% and 80\%. We attribute this difference to the intrinsic properties of the objects. Unlike \ac{1d} objects, \ac{2d} objects have fabric parts that dangle outside the fingertips (\Cref{fig:teaser}). Due to gravity, they are more likely to slide out through the front opening of the gripper. This is evident from the object-dropping rate, which was zero for \ac{1d} objects. Furthermore, the towel is larger and heavier than the cloth, making it more susceptible to the effects of gravity. From the perspective of multi-object training, we do not observe any noticeable differences between models trained on multiple datasets and those trained on a single dataset.

\subsection{Experiments on Unseen Objects}
We evaluate the generalization ability of the learned policy by testing it on previously unseen objects, including an \ac{1d} rope and a \ac{2d} napkin (\Cref{fig:objects}). As shown in \Cref{Tab:3}, although remaining slightly higher for the \ac{1d} object, the success rate on unseen objects decreases to overall 65\% compared with the 80\% for the seen objects. Termination-related failures (early stopping and over-tracing) are also more frequent among the failure modes. This suggests that performance is more affected by differences in the visual appearance of the unseen objects (\Cref{fig:objects}). This interpretation is also supported by the tactile features shown in \Cref{fig:objects_tactile}: the unseen objects (bottom row) exhibit resemblance to the seen objects. Overall, the proposed model adapts to unseen objects but is less accurate in identifying the task endpoint.

\begin{table}
\centering
\caption{Experimental results on unseen objects.}\label{Tab:3}
\setlength{\tabcolsep}{0.8mm}{
\begin{tabular}{c|c|cccc}
\hline
 \textbf{\begin{tabular}[c]{@{}c@{}}Unseen \\ objects\end{tabular}} & \textbf{\begin{tabular}[c]{@{}c@{}}Success rate\\ (Wilson 95\% CI)\end{tabular}} & \textbf{\begin{tabular}[c]{@{}c@{}}Robot \\ collision\end{tabular}} & \textbf{\begin{tabular}[c]{@{}c@{}}Early \\ stopping\end{tabular}} & \textbf{\begin{tabular}[c]{@{}c@{}}Over- \\ tracing\end{tabular}} & \textbf{\begin{tabular}[c]{@{}c@{}}Object \\dropping\end{tabular}}\\ \hline
\textbf{Rope}    & 70.0\% [48.1, 85.5] & 0/20 & 4/20 & 0/20 & 2/20 \\
\textbf{Napkin}  & 60.0\% [38.7, 78.1] & 2/20 & 0/20 & 4/20 & 2/20 \\ \hline
    \end{tabular}}
\end{table}

\section{CONCLUSIONS}
\label{sec:conclusions}
In this work, we present an \ac{il}-based approach for deformable object tracing using a single unified policy with both visual and tactile sensing, implemented in a teleoperation system with multi-modal feedback. To capture both local corrections and global task progression, we introduce a center loss and a task loss. Trained on demonstrations from four \ac{1d} and \ac{2d} deformable objects, the model achieves an overall success rate of 80\%, and ablation studies validate the contribution of each component. Tests on two unseen objects further demonstrate generalizability, achieving a 65\% success rate.

\highlight{Several directions could further improve the current method: (1) combining mechanical intelligence (e.g., specially designed grippers~\cite{zhou2021design, zhou2025hand} such as V-shaped~\cite{yu2024hand} or hole-shaped~\cite{zhu2019robotic}) with machine learning; (2) incorporating more deeply coupled sensor-fusion strategies in policy training~\cite{ou2025pl, wu2025convitac}; (3) expanding the evaluation with more trials and a larger set of objects; and (4) leveraging the modularity of the proposed components, including the visual–tactile teleoperation system, the local center loss, and the goal task loss, which can be readily integrated into other imitation learning algorithms (e.g., Diffusion Policy~\cite{chi2023diffusion}) and further validated for general applicability.}








\bibliographystyle{IEEEtran}
\bibliography{reference}

\end{document}